%% file: main.tex
\documentclass[letterpaper, 10 pt, conference]{ieeeconf}  % Comment this line out if you need a4paper

\IEEEoverridecommandlockouts                              % This command is only needed if 
\usepackage{cite}
\usepackage{amsmath,amssymb,amsfonts}
\usepackage{graphicx}
\usepackage{xcolor}
    
\usepackage[utf8]{inputenc} % allow utf-8 input
\usepackage[T1]{fontenc}    % use 8-bit T1 fonts
\usepackage[bookmarksopen,bookmarksdepth=2]{hyperref}       % hyperlinks
\usepackage{url}            % simple URL typesetting
\usepackage{booktabs}       % professional-quality tables
\usepackage{amsfonts}       % blackboard math symbols
\usepackage{nicefrac}       % compact symbols for 1/2, etc.
\usepackage{microtype}      % microtypography
\usepackage{xcolor}         % colors
\usepackage{siunitx}
\usepackage{algorithm}
\usepackage[noend]{algpseudocode}
\usepackage{amsmath,bm}
\usepackage{mathtools}
\usepackage{centernot}

\usepackage{comment}
\usepackage{cite}
\usepackage{amsmath,amssymb,amsfonts}
\usepackage{graphicx}
\usepackage{standalone}
\usepackage{textcomp}
\usepackage{xcolor}
\usepackage{tikz}
\usepackage{multirow}
\usepackage{tabularx, colortbl}
\usetikzlibrary{automata, positioning, arrows}
\usepackage{siunitx}
\usepackage{graphicx}

\DeclareMathOperator*{\argmax}{arg\,max}

\newtheorem{proposition}{Proposition}

\allowdisplaybreaks

\begin{document}

\title{
Optimal Collaborative Transportation for Under-Capacitated Vehicle Routing Problems using Aerial Drone Swarms
}

\author{Akash Kopparam Sreedhara$^*$, Deepesh Padala$^*$, Shashank Mahesh$^*$, Kai Cui, Mengguang Li, Heinz Koeppl
\thanks{$^*$Authors contributed equally.}
\thanks{This work has been funded by the LOEWE initiative (Hesse, Germany) within the emergenCITY center. The authors are with the Departments of Electrical Engineering and Information Technology, Technische Universität Darmstadt, 64287 Darmstadt, Germany. (e-mail: {\tt\small  heinz.koeppl@tu-darmstadt.de}).}}

\maketitle

\begin{abstract}
Swarms of aerial drones have recently been considered for last-mile deliveries in urban logistics or automated construction. At the same time, collaborative transportation of payloads by multiple drones is another important area of recent research. However, efficient coordination algorithms for collaborative transportation of many payloads by many drones remain to be considered. In this work, we formulate the collaborative transportation of payloads by a swarm of drones as a novel, under-capacitated generalization of vehicle routing problems (VRP), which may also be of separate interest. In contrast to standard VRP and capacitated VRP, we must additionally consider waiting times for payloads lifted cooperatively by multiple drones, and the corresponding coordination. Algorithmically, we provide a solution encoding that avoids deadlocks and formulate an appropriate alternating minimization scheme to solve the problem. On the hardware side, we integrate our algorithms with collision avoidance and drone controllers. The approach and the impact of the system integration are successfully verified empirically, both on a swarm of real nano-quadcopters and for large swarms in simulation. Overall, we provide a framework for collaborative transportation with aerial drone swarms, that uses only as many drones as necessary for the transportation of any single payload.
\end{abstract}

% \begin{IEEEkeywords}
% UAV swarms, collaborative transportation, combinatorial optimization, vehicle routing problem
% \end{IEEEkeywords}

\input{contents/introduction}
\input{contents/related_works}
\input{contents/mtsp_formulation_new}

\input{contents/solutions/solutions}
\input{contents/overview}

\input{contents/experiments}
\input{contents/discussion}

\bibliographystyle{IEEEtran}
\bibliography{references}

\end{document}

%% file: contents/introduction.tex
\section{Introduction}
In recent work, drones have been theorized to provide various civilian services. Possible use cases include search and rescue\cite{drone_search_and_rescue}, performing inspections in hard-to-reach places\cite{power_line_inspection} or last mile deliveries\cite{last_mile_deliveries}, see also
\cite{drone_delivery_interest} for benefits of using drones to achieve the above. 
Drone deliveries have become increasingly sought-after in recent works relating to city logistics\cite{drone_delivery_interest}. This interest has been further accentuated by Google's Project Wing and Amazon's use of quadcopters for their deliveries. A considerable portion of road traffic is caused by package delivery partners, which can be alleviated by using drones to deliver these packages. 
However, due to the limitations of current battery technology, the thrust-to-weight ratio of drones is relatively low\cite{drone_constraints}. This limitation requires larger drones, which are very expensive. Instead, this work explores using multiple under-capacitated drones to deliver heavier payloads cooperatively. Using a swarm of (possibly homogeneous) drones working together to achieve a common goal makes transportation faster, easier to scale, and more efficient\cite{swarm_advantages}.
However, effectively coordinating many drones remains a challenge that will be addressed in this work. Our goal is to provide an integrated solution for collaborative transportation that optimizes transportation plans, using collision-free path planning for coordination.

\input{contents/images/cmtsp_intro}

Fig.~\ref{fig:cmtsp_repr} visualizes our setting in a city logistics scenario. Each drone has a certain capacity and velocity, and each payload has a variable mass. 
The delivery of all payloads is formulated as a list of missions to be completed. 
A single drone can complete some missions, while others require cooperation between drones due to a heavier payload. 
From an economic application point of view, the problem statement requires (i) the minimization of the total time it takes for all drones to perform missions and arrive back at their depot, and (ii) the minimization of total distance travelled, as a proxy for energy or transportation costs. Reducing the time between missions pushes the agents to collaborate, avoiding cases where few agents complete all missions. This results in a complex combinatorial optimization problem, where an improper assignment of drones to missions may lead to underutilization of drones, while an improper ordering of missions may cause drones to travel too far, wait too long for other drones to arrive, or even causes deadlocks.

The above scenario is formulated as a novel multi-objective optimization problem that generalizes the Vehicle Routing Problem (VRP), which in itself generalizes the Multiple Traveling Salesmen Problem (mTSP) and has many applications in fleet management systems or transportation of people \cite{Application_VRP}. 
Our under-capacitated VRP (uVRP) aims at optimal routes for a group of drones to complete a set of joint pickup and delivery missions across different locations, i.e. objects may require more than a single drone to be lifted.
Qualitatively, uVRP introduces timing constraints as a major difference to standard and capacitated VRPs \cite{ralphs2003capacitated}, since it is no longer sufficient only to consider the distances travelled between each mission, but instead one must also consider the time spent waiting for other agents to arrive for a collaborative transportation.
Our setting generalizes VRP and is similarly not only relevant for logistics, but of independent interest to, e.g., moving services, automated construction \cite{augugliaro2014flight}, humanitarian supply chains \cite{HumanitarianSupplyChain} or flood rescue operations \cite{Flooding}. 

Our primary contributions include (i) formulating the first under-capacitated VRP with optimization objectives of interest; (ii) providing an alternating minimization scheme by encoding the solution in a deadlock-free manner, and solving uVRP in practice using metaheuristics; and (iii) demonstrating the integration of collaborative transportation with collision avoidance both in simulation and on a system of real nano quadcopters with payloads in a controlled environment.

%% file: contents/images/cmtsp_intro.tex
\begin{figure}[tb]
\begin{center}
\includegraphics[width=0.8\linewidth]{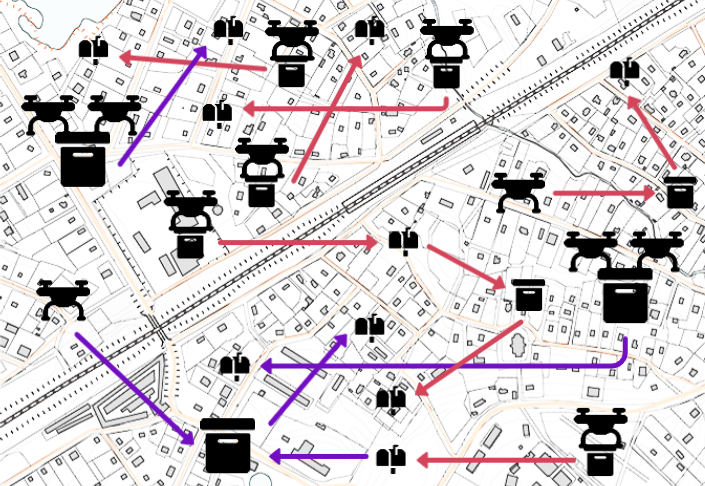}
\end{center}
\caption{\textbf{Drone swarms in a logistics scenario.} Swarms of drones are sequentially allocated to collaborative transportation missions for many packages. Some packages can be transported by single drones (red), while others require multiple drones to transport collaboratively (purple).} % The dotted lines represent ongoing missions, whereas the solid lines represent drones going to pick up a package to start a mission.
\label{fig:cmtsp_repr}
\end{figure}

%% file: contents/related_works.tex
\section{Related Work}
A concept of aerial highways with shared air spaces was introduced in \cite{drone_airways}. However, using single drones for delivering payloads is limiting due to the low thrust-payload ratio of drones, which calls for the swarming of drones to enhance efficiency. Previous works showed how swarms of drones could be used to set up communication networks in areas where communications architecture got damaged due to disasters\cite{data_ferrying}, and how they can help in fire suppression activities\cite{fire_suppression} or surveillance and weather monitoring \cite{weather_monitoring}. The idea of collaboration between decentralized micro-UAVs to transport fabric was introduced in \cite{fabric_transportation}. Load sharing and energy distribution between 2 UAVs to transport objects was explored in \cite{load_sharing}. Further, \cite{control_and_obs_avoidance} proposed control systems for collaborative transport of rigidly attached payloads using multiple UAVs. It also facilitates collision avoidance for stationary obstacles by modelling them as constraints. Control for collaborative transportation still remains subject of active research \cite{ping2023collaborative}. See also, e.g., \cite{schiano2022reconfigurable} for a recent overview. In this work, we enable dynamic collision avoidance for moving obstacles, i.e. other groups of drones.

For coordination, e.g., data ferrying using swarms of drones was proposed in \cite{mtsp_swarm} by formulating the problem as an mTSP, a relaxation of the VRP for drone delivery \cite{drone_vrp}. The proposed mTSP was solved using a genetic algorithm. Variants of TSP are a recurring theme in UAV motion planning\cite{tsp_drone}. Due to their NP-hardness, various metaheuristic algorithms have been employed to solve mTSP and its variants\cite{mtsp_ga, sa_comparison, ctsp_ga}, while \cite{hu2020reinforcement} employ graph neural networks and reinforcement learning to solve mTSP variants. In the context of drone deliveries, the problems are a form of asymmetric mTSP, where each location represents both picking up and delivering a package at once. 
% The cost of travelling from location A to B is not equal to that of travelling from location B to A. Such a generalization comes from the vehicle routing problem (VRP). 
In a capacitated setting, \cite{multi_payload_vrp} further generalizes VRP by introducing drones with the capability to carry multiple packages. A comprehensive overview of the extensive prior literature is found in \cite{cheikhrouhou2021comprehensive}. In contrast to existing work, we introduce under-capacitated VRP, where the VRP is generalized to allow for drones with payload capacities that are lower than the mass of some payloads, requiring additional coordination in-between drones. %This would require facilitation of drones coordinating to deliver packages to their destination.

%% file: contents/mtsp_formulation_new.tex
\section{Under-Capacitated Vehicle Routing Problem}
As discussed in the introduction, the scenario includes $n$ drones $\mathcal D \coloneqq \{ 1, \ldots, n \}$ and $m$ transportation missions $\mathcal M \coloneqq \{ n+1, \ldots, n+m \}$. For a compact Euclidean area of operations $\mathcal X \subseteq \mathbb R^2$, each drone $d \in \mathcal D$ has its depot at location $x_d \in \mathcal X$, while each mission $m \in \mathcal M$ consists of an object with weight $w_m \in \mathbb R$ at $x_m \in \mathcal X$, which is to be transported to destination $y_m \in \mathcal X$. For simplicity, assume homogeneous drones with payload capacity $C \in \mathbb R$ and that each drone can transport at most a single object. Extending to heterogeneous drones or higher capacities is relatively straightforward, but omitted for simplicity of exposition. Then, the number of drones required for mission $m$ is $c_m \coloneqq \lceil \frac{w_m}{C} \rceil$.

In traditional VRPs and mTSPs, the standard MinSum and MinMax objectives correspond to minimizing distance and time taken, which can be interpreted as energy and opportunity costs respectively. However, in under-capacitated VRP there is a pronounced difference. In contrast to simply optimizing either the sum or maximum of travelled distances for each drone, the objectives corresponding to energy and time costs do not sum or maximize the same terms. Instead, while the MinSum formulation still minimizes the sum of travelled distances, the MinMax formulation should instead minimize the maximum time taken over all drones for all missions to complete, which includes not only travelling time, but also time spent waiting until all other required drones arrive at a collaborative transportation mission.

\paragraph{Travelled distance optimization}
To formulate the under-capacitated VRP as an optimization problem, define the set of all drone assignments $\mathcal G \coloneqq \mathcal D \cup \mathcal M$. Then, let $\mathbf t \coloneqq (t^d_{ij})_{d \in \mathcal D, i,j \in \mathcal G}$ be a tuple of binary variables $t^d_{ij}$, equal one whenever drone $d$ travels from assignment $i$ to $j$ as part of the solution $\mathbf t$. We then define distances between depots and missions $i,j \in \mathcal G$ as
\begin{align}
    d_{ij} \coloneqq \begin{cases}
        \lVert x_i - x_j \rVert \quad \text{if} \quad i \in \mathcal D, \\
        \lVert x_i - y_i \rVert + \lVert y_i - x_j \rVert \quad \text{else.} \\
    \end{cases}
\end{align}

Then, considering for now only the distances travelled by each drone, a possible formulation of the problem is
\begin{subequations} \label{eq:opt}
    \begin{alignat}{2}
    & \max_{\mathbf t, \mathbf u} &\quad& \sum_{d \in \mathcal D} \sum_{i,j \in \mathcal G} d_{ij} t^d_{ij} \eqqcolon J_{d} \\
    &\quad \text{s.t.} & & \sum_{d \in \mathcal D} \sum_{j \in \mathcal G} t^d_{ij} = c_i \quad \forall i \in \mathcal M, \label{eq:mnum}\\
    &                  & & \sum_{j \in \mathcal G} t^d_{ij} = \sum_{j \in \mathcal G} t^d_{ji} \leq 1 \quad \forall d \in \mathcal D, i \in \mathcal M, \label{eq:dnum}\\
    &                  & & \sum_{j \in \mathcal G} t^d_{dj} = \sum_{j \in \mathcal G} t^d_{jd} = 1 \quad \forall d \in \mathcal D, \label{eq:depot}\\
    &                  & & u_i - u_j \geq t^d_{ij} - 1 \quad \forall d \in \mathcal D, i \in \mathcal M, j \in \mathcal G, \label{eq:pot}\\
    &                  & & t^d_{ij} \in \{ 0, 1 \}, u_i \in [0, 1] \quad \forall i, j \in \mathcal G, d \in \mathcal D.
    \end{alignat}
\end{subequations}

We note that the above objective is exactly equal to the sum of travelled distances by all drones. Here, constraint \eqref{eq:mnum} ensures that exactly $c_i$ drones perform any mission $i$, constraint \eqref{eq:dnum} ensures that each drone entering a mission must also exit the mission (at most once), and constraint $\eqref{eq:depot}$ ensures that each drone starts and ends at its depot. Finally and most importantly, the constraint \eqref{eq:pot} introduces node potentials $\mathbf u = (u_i)_{i \in \mathcal G}$ as a subtour and deadlock elimination constraint. In other words, it ensures that there is a strict partial order of all missions, according to which the missions can be performed without deadlock: Consider an example with two missions requiring two drones each. If two drones serve both missions but in the opposite order, the drones will forever wait for the other drone, resulting in a deadlock. The existence of node potentials $\mathbf u$ ensures that missions can be executed in order of increasing $u_i$.

\paragraph{Total time optimization}
Based on the above formulation, optimizing instead the total time to finish all missions and return to depots, leads to the objective
\begin{align}
    J_{T} \coloneqq \max_{d \in \mathcal D} \left\{ \sum_{i,j \in \mathcal G} \left( \frac{d_{ij}}{v} + \Delta T^d_{i} \right) t^d_{ij} \right\}
\end{align}
assuming for simplicity a constant drone velocity $v$, where one additionally minimizes waiting times at missions $i$,
\begin{align}
    \Delta T^d_{i} = \max_{d' \in \mathcal D \colon \sum_j t^{d'}_{ij} = 1} T^{d'}_i - T^d_i
\end{align}
until all other drones $d'$ required for mission $i$ arrive, where $T^d_i$ is the time it takes for drone $d$ to reach node $i$, recursively computed by including previous waiting times.

\paragraph{Multi-objective optimization}
For achieving an economic trade-off, we propose a multi-objective optimization via scalarization \cite{ehrgott2006discussion}, i.e. we optimize the weighted sum
\begin{equation} \label{eq:obj}
    J = \mu J_{d} + (1 - \mu) J_{T}
\end{equation}
under some scalarization parameter $\mu \in [0, 1]$ that trades off energy and time costs depending on the application.

In contrast to capacitated VRP, the under-capacitated VRP adds timing constraints. Instead of simply considering the distances travelled by each agent, one must also consider the time spent waiting for other agents to arrive at an object. This changes the problem qualitatively. As a result, existing algorithmic approaches are inapplicable, and in the following we develop specialized algorithmic solutions.

%% file: contents/solutions/solutions.tex
\section{Algorithmic Solution}
\label{cmtsp_solvers}
In this work, the primary algorithmic contribution is in formulating an alternating minimization scheme under a natural encoding of valid solutions for the uVRP, i.e. solutions without deadlocks. Numerically, since the uVRP is NP-hard as a generalization of TSP \cite{cheikhrouhou2021comprehensive}, in this work we employ metaheuristics. Unfortunately, due to the potential for deadlocks, it is not possible to decompose uVRP into independent sub-instances of TSP, in contrast to e.g. mTSP \cite{hu2020reinforcement}, which is a special case of uVRP. Therefore, we design a novel algorithm based on a decomposition of the solution into two parts, in turn avoiding any deadlocks.

\subsection{Alternating minimization algorithm}
The solutions generated by algorithms will be of the form as shown in Table~\ref{table:solution_repr}. Here, $\pi$ is the order in which the payloads are delivered, and $\Psi$ encodes which drones are used for each payload. More precisely, consider permutations $\pi \in \mathrm{Sym}(\mathcal M)$ of missions $\mathcal M$, which define the mission execution order $\pi(1), \ldots, \pi(m)$ and realize potentials in \eqref{eq:opt} by $u_i = \pi(i) / m$. Then, $\Psi \in \mathcal P(\mathcal D)^{\mathcal M}$ defines which drones $\Psi(i)$ will perform missions $i$, where $\mathcal P$ denotes the power set. The assignments $\Psi$ are implemented by $\{0,1\}$-valued matrices as seen in Table~\ref{table:solution_repr}. Noting that the objective \eqref{eq:obj} can be written as a function $J = J(\pi, \Psi)$ of $(\pi, \Psi)$, a tuple $(\pi, \Psi)$ therefore constitutes a solution to the uVRP that avoids deadlocks. 

This decomposition allows us to avoid deadlocks and establishes an algorithm by alternating minimization of $\pi$ and $\Psi$ respectively, i.e. starting with some initial solution, in each iteration $q = 0, 1, \ldots$ of the alternating minimization (AM) algorithm, we compute
\begin{align}
    \Psi^{(q+1)} &\coloneqq \argmax_{\Psi \in \mathcal P(\mathcal D)^{\mathcal M}} J(\pi^{(q)}, \Psi), \label{eq:am1}\\
    \pi^{(q+1)} &\coloneqq \argmax_{\pi \in \mathrm{Sym}(\mathcal M)} J(\pi, \Psi^{(q+1)}). \label{eq:am2}
\end{align}

In general, the AM algorithm converges to a coordinate-wise optimal solution by monotone convergence and $J \geq 0$.
\begin{proposition}
    The AM algorithm given by \eqref{eq:am1}-\eqref{eq:am2} converges in $J(\pi^{(q)}, \Psi^{(q)})$ as $q \to \infty$.
\end{proposition}

\input{contents/solutions/solution_representation}

\input{contents/solutions/heuristics}

%% file: contents/solutions/solution_representation.tex
\begin{table}[tb]
\centering
\caption{Solution representation by permutations $\pi$ \& assignments $\Psi$.} % This is also the encoding technique which is going to be used by the algorithms which will optimize $\pi$ and $\Psi$
\begin{tabular}{c c c c c c}
\hline

\multirow{2}{*}{\begin{tabular}{c}
    Mission \\ Order ($\pi$)
\end{tabular}} & \multicolumn{5}{c}{Drone Assignments ($\Psi$)} \\
\cline{2-6}
& Drone 1 & Drone 2 & Drone 3 & $\dots$ & Drone $n$ \\
\hline
1: Payload 4 & 1 & 0 & 1 & $\dots$ & 0 \\
%\hline
2: Payload 2 & 0 & 1 & 0 & $\dots$ & 1 \\
%\hline
3: Payload 7 & 1 & 1 & 0 & $\dots$ & 0 \\
%\hline
$\vdots$ & $\vdots$ & $\vdots$ & $\vdots$ & $\ddots$ & $\vdots$ \\
%\hline
$m$: Payload \dots & 0 & 0 & 1 & $\dots$ & 1 \\
\hline
\end{tabular}
\label{table:solution_repr}
\end{table}

%% file: contents/solutions/heuristics.tex
\subsection{Metaheuristics}
\label{heuristic_solver}
Since the optimization of each subproblem \eqref{eq:am1} and \eqref{eq:am2} remains difficult, the alternating minimization algorithm instantiates two metaheuristic algorithms, $\rho_\Psi$ and $\rho_\pi$, for the combinatorial optimization problems \eqref{eq:am1} and \eqref{eq:am2}.
Simulated annealing can be better than Genetic Algorithms for TSP\cite{sa_comparison}, while genetic algorithms are widely used to solve mTSP\cite{mtsp_ga}. Combining both algorithms has been proven effective also in other variants of the mTSP\cite{ctsp_ga}. 
Hence, in this work, the SAGA algorithm in Algorithm~\ref{alg:am} implements the alternating optimization scheme by combining a genetic algorithm (GA) $\rho_\Psi$ and a simulated annealing (SA) algorithm $\rho_\pi$. The SA algorithm $\rho_\pi$ is tasked with finding the optimal assignment of payloads to each mission. The GA $\rho_\Psi$ finds the optimal assignment of drones for each mission. Depending on the requirements and constraints, the optimal fitness/cost function can be chosen by adjusting the trade-off between the total time to complete all missions and the total distance travelled by all drones combined. See also Section~\ref{experiments}.

\begin{algorithm}[tb]
\caption{SAGA for uVRP}
\label{alg:am}
\begin{algorithmic}[1]
        \State Randomly generate initial $\pi$.
        \State Randomly generate an initial population of drone assignments $P = \{ \Psi_1, \ldots, \Psi_{P_{\mathrm{GA}}} \}$.
        \For {iterations $n=1, \ldots, N_{\mathrm{alt}}$}
            \State Optimize $P = \rho_\Psi(\pi)$ using GA $\rho_\Psi$; $N_G$ iterations
            \State Optimize $\pi = \rho_\pi(P)$ using SA $\rho_\pi$; $N_S$ iterations
        \EndFor \\
        \Return $\pi$ and $\argmax_{\Psi_i \in P} J(\pi, \Psi_i)$ with best cost.

\end{algorithmic}
\end{algorithm}

\paragraph{Genetic Algorithm}
Genetic algorithms work by initialising a random population and scoring it. Multiple iterations of mutation and crossover are performed, followed by selection and reinsertion in the population based on the fitness of each genotype $\Psi_1, \ldots, \Psi_{P_{\mathrm{GA}}}$ (individuals in the population $P$). This ensures that the population is optimized over its iterations. The encoding for the uVRP solution genotype is as in Table~\ref{table:solution_repr}, assuming homogeneous drones.

For initialization, the population is chosen such that each genotype is a valid solution. Drone assignments for each payload $m$ are obtained by sampling $c_m$ indices from $\{1, \ldots, n\}$, and using those drones for the payload, where we recall that $c_m$ is the number of drones required for mission $m$ of the selected payload, and $n$ is the number of drones.
The two-parent uniform crossover\cite{uniform_crossover} is used as depicted by Fig.~\ref{fig:uniform_crossbreeding}, with a mutation operator that randomly replaces the drones in use with the same number of randomly sampled drones to ensure that the drones can carry the payload. The mutation probability was tuned to get the best outcome. The GA uses elitist reinsertion and roulette wheel selection to ensure optimal population generation.

\input{contents/solutions/table_crossover}

\paragraph{Simulated Annealing}
The Simulated Annealing (SA) algorithm has been shown to be effective in combinatorial optimization problems where permutations have to be optimized. It works by simulating the heat treatment process of annealing, which alters the properties of a material to make it more suitable to work on, by heating the material to a certain temperature and cooling it to room temperature. 
As the material (solution) cools down, it obtains more desirable properties (better cost $J$). Starting with a very high temperature, the SA simulates cooling with a cooling rate and works by transfiguring the payload order by performing one of the following randomly sampled mutation techniques:
\begin{itemize}
    \item {Swap Mutation:} Two payloads are chosen randomly in the mission order and swapped.
    \item {Reverse Mutation:} Two indices which are $m/2$ apart are chosen randomly, and the slice between them is reversed. 
    \item {Insertion Mutation:} A payload is randomly removed from $\pi$ and inserted into a different index.
    \item {Insert Slice Mutation:} A random slice is chosen, removed from $\pi$ and inserted into a random index.
\end{itemize}

SA avoids converging at a local optimum by using the Metropolis criterion to select a solution with a certain probability even if it is worse than the current solution. At higher temperatures, the SA explores the solution space more, using the probability of accepting a worse solution $(\pi_p, \Psi_p)$,
\begin{equation}
    M_P = \min \left\{1, \exp \left( -\frac{J(\pi_p, \Psi_p) - J(\pi_c, \Psi_c)}{T} \right) \right\}
\label{eqn:metropolis}
\end{equation}
when $\pi_c$ and $\Psi_c$ are the current solution, and $T$ is the temperature, which decays by multiplying with $\gamma \in (0, 1)$ at each step. Here, $\Psi_p$ and $\Psi_c$ are selected as the best (three) assignments for $\pi_p$ and $\pi_c$ respectively from current population $P$ (and averaging the three costs). %This helps the solution approach the global optimum as it cools down completely.

It is important to note that any changes by GA to $\Psi$ in matrix form (in Table~\ref{table:solution_repr}) will not affect $\pi$ as long as the drones assigned to the payloads can transport them. However, any changes to the permutation in $\pi$ without corresponding changes to $\Psi$ might render the solution invalid. To prevent this, any transfiguration applied to $\pi$ by $\rho_\pi$ is also applied to $\Psi$, effectively ensuring that changing the payload order does not change which drones will carry it. In other words, mutation operations applied to the payload order are also applied to all corresponding indices on each genome in the GA. The temperature is reset to the original starting temperature for each new alternating iteration, since each iteration of alternating minimization fully minimizes in each step. Table~\ref{table:vars} shows the parameters used in this work.

\begin{table}[b]
\centering
\caption{Parameters of SAGA}
\begin{tabularx}{0.75\linewidth}{l c}
\toprule
    \textbf{Algorithm Parameters} & \textbf{Value} \\ \midrule
    GA iterations ($N_G$) & 200 \\
    SA iterations ($N_S$) & 500 \\
    Alternating optimizer iterations ($N_\mathrm{alt}$) & 3 \\
    \midrule
    Population size ($P_{\mathrm{GA}}$) & 50 \\
    Individuals per parents & 3 \\
    Selection ratio & 0.8\\
    Mutation rate & 0.05 \\
    Reinsertion ratio & 0.7 \\
    \midrule
    SA cooling rate ($\gamma$) & 0.97 \\
    SA starting temperature & 15000 \\
    \bottomrule
\end{tabularx}
\label{table:vars}
\end{table}

%% file: contents/solutions/table_crossover.tex
\tikzstyle{c-rectangle2} = [rectangle, ultra thick, rounded corners, minimum height=1cm,text-centered, text width = 3cm, draw=black, fill=white]
\tikzstyle{arrow} = [semithick, ->, shorten > = -2pt, shorten < = -2pt]

\begin{figure}[b]
\centering
\begin{tikzpicture}[node distance = 2cm]
\node (parent_1) [shape=rectangle,xshift = -5.5cm , scale=0.75, draw] {
\begin{tabular}{c c c}
  \colorbox{white}{Drone 1} & \colorbox{white}{Drone 2} &\colorbox{white}{Drone 3} \\
  \hline
  \colorbox{white}{0} & \colorbox{white}{1}  &\colorbox{white}{0} \\
  \colorbox{white}{1} & \colorbox{white}{1} & \colorbox{white}{0} \\
  \colorbox{white}{0} & \colorbox{white}{1} & \colorbox{white}{1} \\
  \colorbox{white}{1} & \colorbox{white}{0} & \colorbox{white}{0} \\
  \hline
  
  \multicolumn{3}{c}{parent 1}
  \end{tabular}
};
\node(parent_2) [shape = rectangle, xshift=-1cm,scale = 0.75 ,draw] {
\begin{tabular}{c c c}
  {Drone 1} & {Drone 2} &{Drone 3} \\
  \hline
  \rowcolor{lightgray}
  \colorbox{lightgray}{1} & \colorbox{lightgray}{0}  &\colorbox{lightgray}{0} \\
  
  \rowcolor{lightgray}
  \colorbox{lightgray}{1} & \colorbox{lightgray}{0} & \colorbox{lightgray}{1} \\
  \rowcolor{lightgray}
  \colorbox{lightgray}{1} & \colorbox{lightgray}{1} & \colorbox{lightgray}{0} \\
  \rowcolor{lightgray}
  \colorbox{lightgray}{0} & \colorbox{lightgray}{0} & \colorbox{lightgray}{1} \\
  \hline
  \multicolumn{3}{c}{parent 2}
  \end{tabular}
};

\node(offspring) [shape = rectangle, xshift=-3.2cm,scale = 0.75, yshift=-4cm,draw] {
\begin{tabular}{c c c}

  \colorbox{white}{Drone 1} & \colorbox{white}{Drone 2} &\colorbox{white}{Drone 3} \\
  \hline
  \colorbox{white}{0} & \colorbox{white}{1}  &\colorbox{white}{0} \\
    \rowcolor{lightgray} 
   1 & 0 & 1 \\
  \colorbox{white}{0} & \colorbox{white}{1} & \colorbox{white}{1} \\

    \rowcolor{lightgray}
  0 & 0 & 1 \\
  \hline
  \multicolumn{3}{c}{offspring}
  \end{tabular}
};

% \node(mask) [shape = rectangle, xshift=-6.2cm,scale = 0.75, yshift=-4cm,draw] {
% \begin{tabular}{c}

% % \centering
%   \colorbox{white}{mask} \\
%   \hline
%   \colorbox{white}{Parent 1} \\
%   \colorbox{white}{Parent 2} \\
%   \colorbox{white}{Parent 1} \\
%   \colorbox{white}{Parent 2} \\
%   \\
%   \end{tabular}
% };
 \draw[->] (parent_1.south) -- (-3.2,-1.5);
  \draw[->] (parent_2.south) -- (-3.2,-1.5);
  \draw[->] (-3.2,-1.5) -- (offspring.north);
%\draw[arrow]  (parent_1.south) edge (offspring.north west);
%\draw[arrow]   (parent_2.south) edge (offspring.north east);

\end{tikzpicture}
\caption{\textbf{Uniform two-parent crossbreeding.} Shown in the figure are 2 parent $\Psi$s from the GA population creating an offspring $\Psi$. The example problem statement requires the transportation of 4 packages using 3 drones. The drone configuration for each payload is randomly chosen from one of parents 1 and 2, to make it into the offspring.}
\label{fig:uniform_crossbreeding}
\end{figure}
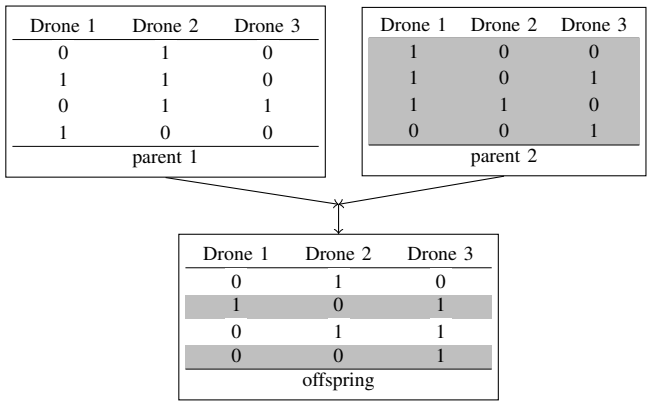

%% file: contents/overview.tex
\section{Implementation and Methodology}
\begin{figure}[tb]
\centering
\includegraphics[width = \linewidth]{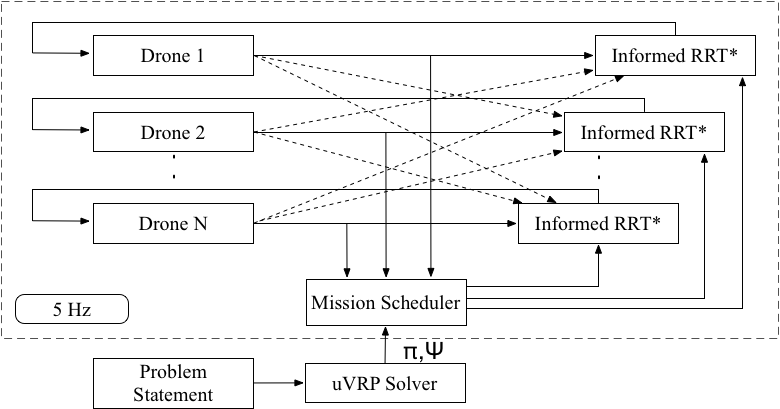}
\caption{
\textbf{Integration of algorithms for swarm delivery.} The uVRP solver generates a near-optimal solution, which is parsed by the mission scheduler to assign drone goals, i.e. pick up, drop off, and transportation. The high-level goals are passed to the Informed RRT* path planner, which plans optimal motion sequences for the drones to execute to complete their missions.
} 
\label{fig:algo_integration}
\end{figure}

Overall, the uVRP algorithm is integrated into a real system as described in the following, by introducing collision-free path planning and drone controllers, see also Fig.~\ref{fig:algo_integration}.

\subsection{Hardware}
This work utilises Crazyflie 2.0 nano-drones equipped with a Lighthouse positioning system (Steam VR Base Station 2.0). Communicating positions and velocities of drones between the central PC (AMD Ryzen 9-5900HX CPU and 16GB RAM) and each drone is done at 5 \unit{\hertz}. A single payload consists of a 3D-printed block with freely moving, cylindrical permanent magnets that are attracted to the iron plate on the payload docker of drones. We attach multiple payloads using carbon fibre rods for heavier and larger, collaboratively transported payloads. The drop-off point employs a stronger magnet that detaches the payload via a stronger magnetic force than the force between the metallic plate and the payload, facilitating successful payload drop-offs. See also Fig.~\ref{fig:catcheye} for images.

\subsection{Control and Mission Tracking} 
\label{control_and_scheduling}
The uVRP solver assigns a group of drones to each payload, along with the order in which the payloads must be delivered, as shown in Table~\ref{table:solution_repr}. The drone group may consist of only a single drone, provided that the drone is capable of transporting the payload. When multiple drones are in a group, the radius is increased to a circle, enclosing all drones in the group. A single thread is used to plan a path for the group.
A ROS2 node/topic framework is used to integrate all the subsystems. The path planner receives information about the radius, velocity and the current position of each drone group. The high-level commander assigns set points to each drone to make each group follow the path via a cascaded PID controller, which enables position control of each drone. 

\paragraph{Pick-up mechanism}
A pickup manoeuvre cannot be executed unless all drones arrive at the payload. Upon arrival, drones move synchronously to pick up a payload. Upon completion, the path planner is notified to create a group with updated radius, encompassing all involved drones.

\paragraph{Drop-off mechanism}
The drone group descends synchronously to drop the package. Once done, the group is split into separate groups for each drone, with their corresponding radius and positions being tracked. %All planner threads which were deactivated are then reactivated.

\input{contents/drone_scheduler}
\input{contents/collision_avoidance}

%% file: contents/drone_scheduler.tex
\paragraph{Mission scheduler}
The algorithm to schedule drones iterates over $\pi(1), \ldots, \pi(m)$. The corresponding missions are added to a queue for each drone. The missions to be added to the drones' queues are decided by $\Psi$.  Drones execute a mission when all drones arrive at the payload pickup location. Once payload drop-off is complete, the corresponding mission is removed from the queues of all involved drones. A drone returns to its depot once its mission queue is empty.

%% file: contents/collision_avoidance.tex
\subsection{Path Planning}
\label{path_planning}
A decentralised path planning algorithm based on the Informed RRT*\cite{informedrrtstar} planner has been implemented. The sampling-based path planner on each drone \cite{RRT_uav_collision_avoidance} facilitates collision avoidance, and was tested in a multi-threaded platform where each thread ran on its corresponding drone. 
Introducing a latency of up to $\SI{3.94}{\milli\second}$ (wireless delays) in the information shared between threads did not affect the algorithm's performance. Decentralised communication and embedded systems implementations for the Crazyflie 2.1 drones are out of scope.
The data shared between drones consists of the positions and headings, their velocities, and the radius of each drone group in the swarm. %  ($x_i$, $y_i$, $\theta_i$ - representing the yaw angle) ($\dot{x_i}$, $\dot{y_i}$, $\dot{\theta_i}$)

The SE(2) space was used for the planner to ensure all drones use the same altitude in the airspace to avoid the downwash effect. Extending the algorithm to SE(3) is straightforward and uncomplicated if required. %The SE(2) state is $(x, y, \theta)$.
Sampling-based path planners work by sampling a set of states in the SE(2) workspace, checking their validity and choosing a subset of the valid samples ($V$) for optimal motion planning for the drone. All the sampled states are validated by checking for possible collisions with all active drones. Shadows are projected based on the current velocity of each drone to avoid deadlocks and future collisions. The sampled states which pass the above checks make it into $V$.
Paths are re-planned in case the drones stray too far from the previous path or if the drones' new velocity projections predict collisions.

%% file: contents/experiments.tex
\begin{figure*}
    \centering
    \hfill{}
    \includegraphics[height=3.2cm]{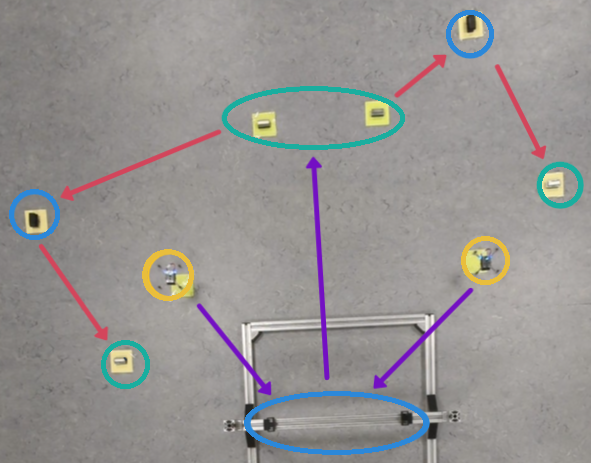}
    \hfill{}
    \includegraphics[height=3.2cm]{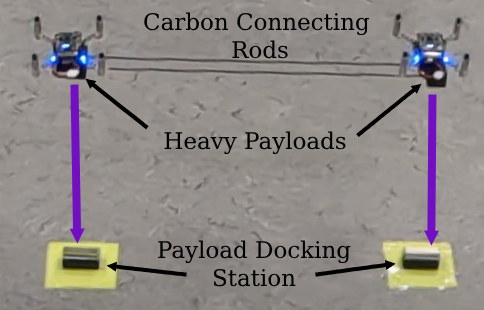}
    \hfill{}
    \includegraphics[height=3.2cm]{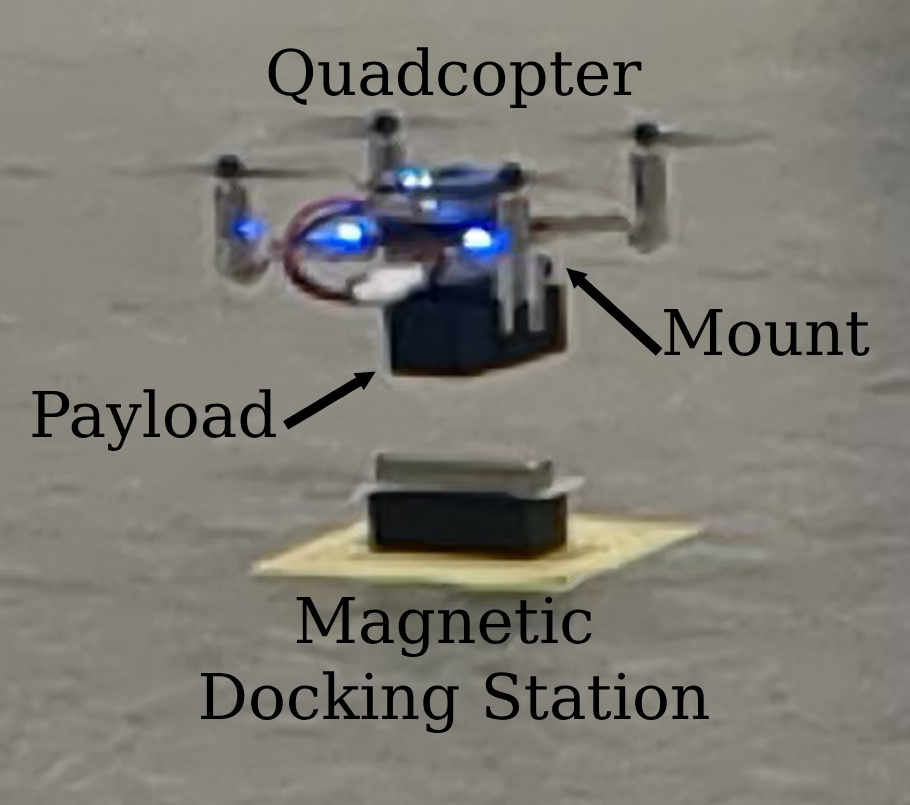}
    \hfill{}
    \includegraphics[height=3.5cm]{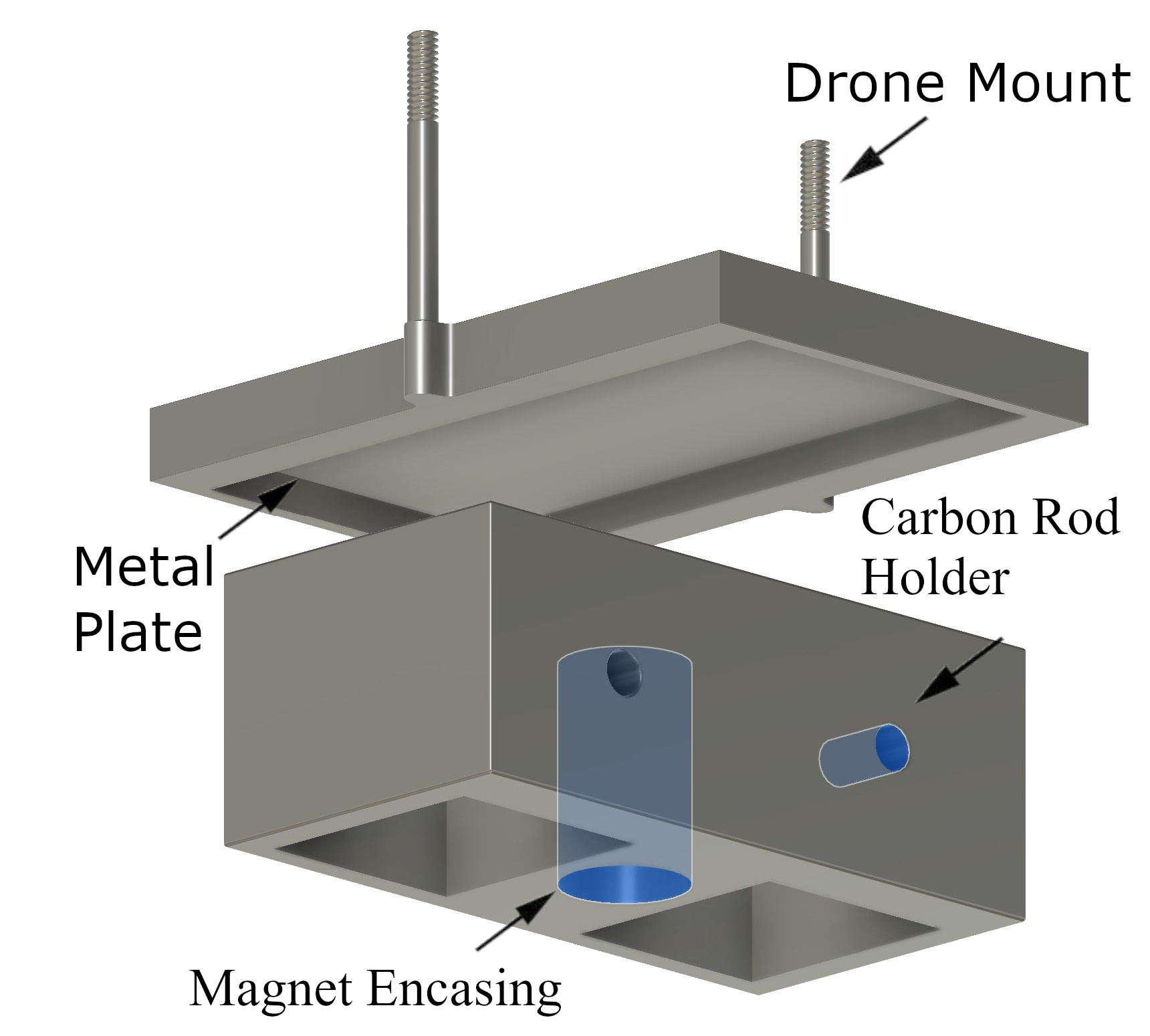}
    \label{fig:real}
    % \vspace{1.0ex}
	\caption{
        \textbf{Real world system integration.}
        The integrated system is successfully verified in practice on real nano quadcopters while avoiding collisions between drones. Left: Example demonstration for two drones that first transport an object collaboratively, and then transport two single drone payloads, colors as in Fig.~\ref{fig:cmtsp_repr}; Right: Hardware setup, including the 3D-printed payloads and docking stations.
    \label{fig:catcheye}}
\end{figure*}

\begin{table}[b!]
\centering
\caption{Comparing SAGA and simulation with path planning on a single problem instance ($\pm$ $95\%$ confidence interval, 50 trials).}
\begin{tabular}{ccccc}
\toprule
\multirow{2}{*}{$n,m$} & \multicolumn{2}{c}{SAGA Predicted Output} &\multicolumn{2}{c}{Actual output}\\
\cline{2-5}
 & Distance(\unit{\metre}) & Time(\unit{\sec}) & Distance(\unit{\metre}) & Time(\unit{\sec})  \\
\hline
3, 15  &302.0 $\pm$ 4.6 &245.7 $\pm$ 5.1 & 377.0 $\pm$ 0.6
& 256.2 $\pm$ 0.9\\
4, 20  &419.6 $\pm$ 4.5 & 295.2 $\pm$ 5.1 & 509.3 $\pm$ 1.0
 & 336.8 $\pm$ 1.7 \\
5, 25  &378.4 $\pm$ 2.6 &202.8 $\pm$ 2.8 & 495.3 $\pm$ 1.1 & 208.1 $\pm$ 2.3\\
5, 30  &465.0 $\pm$ 4.2 &259.3 $\pm$ 5.1 & 665.5 $\pm 2.8$ & 298.2 $\pm$ 5.2 \\
\bottomrule
\end{tabular}
\label{table:am_vs_irl}
\end{table}

\begin{figure}[b!]
    \centering
    \includegraphics[width=\linewidth]{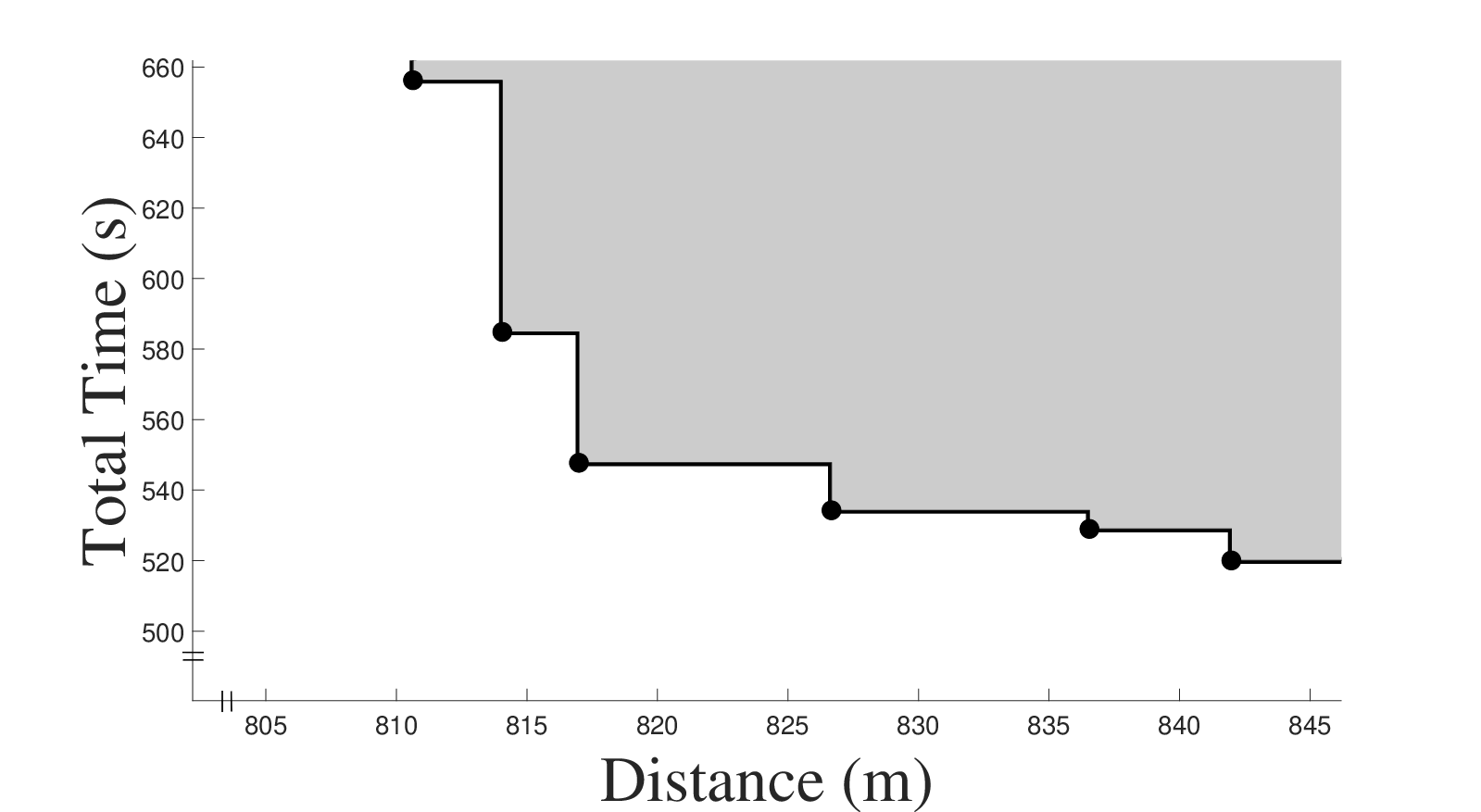}
    \caption{\textbf{Pareto Frontier.} A parameter sweep of $\mu$ was performed in the interval $[0.15, 0.9]$ with increments of $0.15$. For each value of $\mu$, the algorithm was run 800 times. %This illustrates the resulting Pareto frontier. 
}
    \label{fig:pareto}
\end{figure}

\section{Experiments and Results}
\label{experiments}
For evaluation in simulation, problem statements are generated by adding drones and payloads uniformly randomly in a workspace of size $\SI{4}{\metre} \times \SI{4}{\metre}$. The delivery location of the payloads had a randomly generated point with a distance $X$ from the starting point such that $X \sim \mathcal{N}(\SI{2}{\metre},\,\SI{4}{\metre^2})$.

The trade-off between the total distance travelled and time taken is shown as a Pareto frontier in Fig.~\ref{fig:pareto}. The value of $\mu$ can be adjusted according to the curve for a given use case to minimize the economic cost (here, we show the results for $n=4$ drones and $m=100$ payloads).

Table~\ref{table:am_vs_irl} displays the difference in the distance and total time predicted by SAGA and uVRP against the integrated system when simulated with path planning, collision avoidance, using a single integrator model in the bullet physics library and the open motion planning library \cite{ompl}. As seen, the difference is relatively small, i.e. the uVRP model is accurate for practical purposes.

\begin{table}[b!]
    \centering
    \caption{Comparison of SAGA against Random Assignment (RA): Expected performances with $95\%$ confidence interval ($\pm$) averaged over $100$ randomly generated problem instances.}
    \setlength\tabcolsep{3pt}
    \begin{tabularx}{\linewidth}{ccccc}
         \toprule
         $n,m$ & Alg & $J_\mathrm{time}$ (\unit{\sec}) & $J_\mathrm{dist}$ (\unit{\metre}) & $J$ ($\mu = 0.2$) \\
         \toprule
         \multirow{2}{*}{$4, 60$} & RA & 378.48 $\pm$ 24.14 & 506.31 $\pm$ 30.37 & 403.6 $\pm$ 25.4\\
            & SAGA & \textbf{310.08 $\pm$ 9.0} & \textbf{495.46 $\pm$ 11.28} & \textbf{347 $\pm$ 9.5} \\
        \midrule
        \multirow{2}{*}{$5, 60$} & RA & 334.98 $\pm$ 25.39 & 509.19 $\pm$ 31.06 & 369.8 $\pm$ 26.5 \\
            & SAGA & \textbf{242.49 $\pm$ 6.59} & \textbf{491.08 $\pm$ 10.60} & \textbf{292.6 $\pm$ 7.4} \\
            
        \midrule
         \multirow{2}{*}{$4, 100$} & RA & 644.58 $\pm$ 9.21 & 854.42 $\pm$ 10.27 & 686 $\pm$ 9.4 \\
            & SAGA & \textbf{520.03 $\pm$ 9.55} & \textbf{828.08 $\pm$ 9.34} & \textbf{581.6 $\pm$ 9.5} \\
            
        \midrule
         \multirow{2}{*}{$5, 100$} & RA & 573.60 $\pm$ 9.84 & 850.42 $\pm$ 13.52 & 628.4 $\pm$ 10.8\\
         & SAGA & \textbf{411.63 $\pm$ 9.047} & \textbf{829.00 $\pm$ 14.16} & \textbf{494.8 $\pm$ 10.1} \\
        \midrule
          \multirow{2}{*}{$4, 300$} & RA & 2044.24 $\pm$ 22.58 & 2641.69 $\pm$ 27.75 & 2163.4 $\pm$ 23.6\\
         & SAGA & \textbf{1716.57 $\pm$ 29.35} & \textbf{2613.04 $\pm$ 32.59} & \textbf{1895.4 $\pm$ 29.9} \\
          \midrule
          \multirow{2}{*}{$8, 300$} & RA & 1378.08 $\pm$ 17.98 & 2651.36 $\pm$ 29.711 & 1632.6 $\pm$ 20.3 \\
         & SAGA & \textbf{911.13 $\pm$ 18.12} & \textbf{2551.94 $\pm$ 30.14} & \textbf{1422 $\pm$ 20.524} \\
          \midrule
          \multirow{2}{*}{$20, 300$} & RA & 699.15 $\pm$ 11.84 & 2650.66 $\pm$ 34.11 & 1089.2 $\pm$ 16.3\\
         & SAGA & \textbf{515.20 $\pm$ 14.97} & \textbf{2487.90 $\pm$ 39.05} & \textbf{909.7 $\pm$ 19.8} \\
        \bottomrule
        \end{tabularx}
    \label{table:am_vs_random}
\end{table}

In Table~\ref{table:am_vs_random}, the performance of SAGA was compared to random scheduling, i.e. choosing the best solution from the initial population of SAGA. Since to the best of our knowledge, there exist no algorithms for the novel uVRP formulation, comparisons can only be drawn between SAGA and RA. Other solvers available for VRPs or mTSP cannot be compared due to the uVRP's unique necessity of avoiding deadlocks, which the other solvers do not address. It can be seen that the algorithm successfully optimizes the solution beyond random initializations.

Lastly, an exemplary deployment in the real world is shown in Fig.~\ref{fig:catcheye} and the supplementary video\footnote{\url{https://www.youtube.com/watch?v=SAPVUDS24mA}}, where we successfully deploy our results in a real system of nano quadcopters and show the applicability of our developed framework.

%% file: contents/discussion.tex
\section{Discussion and Conclusion}
%%% Recap the results
Overall, we have provided an approach to efficient collaborative transportation via homogeneous drone swarms by formulating a novel extension of VRPs with algorithmic solutions, and successfully providing system integrations for both simulated and real aerial drones.
The framework could be expanded upon in future works: The payloads designed for demonstration are known, uniform masses, alleviating the requirement to design specialized control. However, in practice payloads may be unknown. Online system identification could determine package parameters, and coordination algorithms for partial information could be developed. 
Additionally, when drones collaborate to pick up a package, a single path planning instance is used to plan the path of the entire group. Instead, the state sampling and validation process can be delegated to make motion planning more decentralized and efficient for larger swarms. 
Finally, the uVRP solver is based on metaheuristics. Future works could explore solving the problems by leveraging attention mechanisms in graph neural networks and reinforcement learning, or by applying linear programming relaxations.